\documentclass[10pt,twocolumn,letterpaper]{article}

\usepackage{cvpr}
\usepackage{times}
\usepackage{epsfig}
\usepackage{graphicx}
\usepackage{amsmath}
\usepackage{amssymb}
\usepackage{url}

\usepackage[breaklinks=true,bookmarks=false]{hyperref}

\cvprfinalcopy 

\def\cvprPaperID{****} 

\begin{document}

\title{Progressively Diffused Networks for Semantic Image Segmentation}

\author{Ruimao Zhang $^{1}\footnotemark[1]$,~~Wei Yang $^{2}\footnotemark[1]$,~~Zhanglin Peng $^{3}$,~~Xiaogang Wang $^{2}$,~~Liang Lin $^{1}$ \\
\\
$^{1}$~Sun Yat-sen University, Guangzhou, China\\
$^{2}$~The Chinese University of Hong Kong, Hong Kong, China\\
$^{3}$~SenseTime Group Limited, Shenzhen, China\\
}


\maketitle
\footnotetext[1]{The first two authors contribute equally to this work.}

\begin{abstract}
  This paper introduces Progressively Diffused Networks (PDNs) for unifying multi-scale context modeling with deep feature learning, by taking semantic image segmentation as an exemplar application. Prior neural networks such as ResNet~\cite{C:ResNet} tend to enhance representational power by increasing the depth of architectures and driving the training objective across layers. However, we argue that spatial dependencies in different layers, which generally represent the rich contexts among data elements, are also critical to building deep and discriminative representations. To this end, our PDNs enables to progressively broadcast information over the learned feature maps by inserting a stack of information diffusion layers, each of which exploits multi-dimensional convolutional LSTMs (Long-Short-Term Memory Structures). In each LSTM unit, a special type of atrous filters are designed to capture the short range and long range dependencies from various neighbors to a certain site of the feature map and pass the accumulated information to the next layer. From the extensive experiments on semantic image segmentation benchmarks (e.g., ImageNet Parsing, PASCAL VOC2012 and PASCAL-Part), our framework demonstrates the effectiveness to substantially improve the performances over the popular existing neural network models, and achieves state-of-the-art on ImageNet Parsing for large scale semantic segmentation.

\end{abstract}

\section{Introduction}



In the literature, representation learning~\cite{J:DeepLearning} is a set of methods to automatically discover the representation of raw data for intelligent tasks. Recent developed Deep Convolutional Neural Networks (CNN)  ~\cite{J:CNN,C:ImageNet,C:VeryDeep,C:FCN,C:ResNet} is one of representation learning method with multiple levels of representation, which transforms the raw input image into abstract features by stacking several non-linear modules. To enhance the representation power of such architecture, prior works~\cite{C:VeryDeep,C:GoingDeeper,A:highway,C:ResNet} focus on increasing the depth of architectures and driving the training objective across layers. These methods achieve great success on the challenge ImageNet competition~\cite{C:ImageNet,C:ResNet}, and the semantic segmentation task has also greatly benefited from such ``very deep" model~\cite{C:DeepLab}. However, increasing network depth makes it difficult to design the architecture, while
the gradient vanishing makes the model hard to train.


\begin{figure}[t]
\centering
\includegraphics[width=\linewidth]{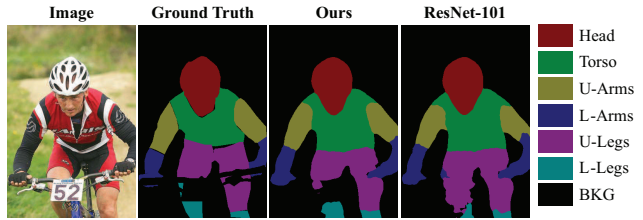}
\caption{An example of semantic image segmentation with and without incorporating spatial dependencies. From left to right are input image, groundtruth labeling, segmentation result by proposed PDNs and segmentation result by ResNet-101~\cite{C:ResNet}. It is obvious that our architecture can capture the rich contexts among pixels and achieve the more reasonable global reasoning.}
\label{fig:example}
\vspace{-1em}
\end{figure}

On the other hand, another branch of works trying to explore rich contexts in the visual data have also received much attention in literature~\cite{C:full-crfs,J:MeanField,C:ContextDriven}. Recent evidence reveals incorporating graphic model, e.g. CRF~\cite{C:DeepLab} or MRF~\cite{C:DPN,A:StructuredNetworks} to the output confidence maps of CNN is of crucial importance, which greatly improves the accuracy of dense prediction. But these context modeling methods require careful pairwise constraints design and do not explicitly enhance the pixel-wise representation, leading suboptimal segmentation results.

An alternative scheme focuses on exploiting Long Short Term Memory (LSTM) networks to automatically learn the spatial dependencies. These data-driven methods use contextual information to enhance intermediate feature representations, achieving promising results on semantic segmentation task~\cite{C:LSTM-RNN,C:SpatialLSTM,C:LG-LSTM,C:Graph-LSTM}, where the property of long-range dependencies is used to pass the information between neighbor pixels layer by layer. However, in terms of information diffusion range in each layer, most existing approaches~\cite{A:Grid-LSTM,C:PixelRNN,C:LG-LSTM,C:Graph-LSTM} have only explored well-designed short distance. A wider range of information diffusion is achieved by stacking multiple LSTM layers. As illustrated in Fig.~\ref{fig:propagation} (a) and Fig.~\ref{fig:propagation} (b), the feature enhancement of each position by above methods is determined by the short distance neighbors (e.g. the closest 2 to 8 adjacent positions) in each layer, limiting the breadth and speed of information diffusion. At the same time, all of the existing methods are based on fully-connected LSTM, and the computational cost of these methods is another obvious limitation.



In this paper, we propose a novel Progressively Diffused Networks (PDNs) that extends the traditional neural network structure from learning rather complex pattern layer by layer to spreading the contextual information in image plane, and demonstrate its superiority on various semantic segmentation tasks. PDNs introduces a stack of information diffusion layers for context modeling, each of which contains several multi-dimensional Long Short-Term Memory (LSTM) networks.  Instead of spreading information to a fixed number of adjacent positions, the diffusion layer allows to propagate information from a certain position to a large range in the image plane. It can effectively expand the scope of communication in a single layer, while improving the speed of information diffusion in multi layers.


Specifically, we propose two types of diffused LSTM, one called spatial LSTM and another called depth LSTM. Each spatial LSTM in the diffusion layer generates a certain type of contextual feature maps. Intuitively, these contextual feature maps have different meanings with the ones generated by traditional convolutional methods. The value in the convolutional maps represents the response of a local area under a certain pattern. In contrast, each site of the contextual feature maps involves the information it will propagate to its neighbors in the next state. Different from spatial LSTM, we incorporate the depth LSTM~\cite{C:LG-LSTM} into diffusion layer to realize the communication of each site from one layer to the next.


In each diffused LSTM unit, the special atrous filters~\cite{A:deeplab2} for each contextual feature maps are used for each position to capture the diverse neighborhood information in a large range of local area. Finally, these filtering results will be integrated to calculate the information of each site passed to its neighbors or to itself in the next layer. Compared with the fully-connected LSTM in previous works~\cite{A:Grid-LSTM,C:SpatialLSTM,C:LG-LSTM,C:Graph-LSTM}, this convolution-based version is more intuitive, and can significantly improve the computational efficiency.


This paper has following four contributions. (1) We propose a novel deep architecture named Progressively Diffused Networks. The stacked diffusion layers allow the contextual information spread from a certain position to a large range on natural image. (2) We design a novel LSTM-based layer, which contains several convolutional LSTM. It can generate a series of contextual feature maps, and the convolutional LSTM in the next layer can exploit these maps to further guide the feature representation of each site. (3) A special type of atrous filters are incorporated into proposed convolutional LSTM, each of which is corresponding to a special contextual feature map. Through the convolutional operation, we reduce the parameter space in context modeling. (4) We obtain state-of-art results on three challenging
datasets, i.e. ImageNet Parsing Dataset~\cite{A:ImageNetParsing}, PASCAL-Part Dataset~\cite{C:Pascal-Part} and PASCAL VOC 2012 semantic segmentation benchmark~\cite{J:PascalVOC}.

\section{Related Work}

\begin{figure*}[t]
\centering
\includegraphics[width=0.9\linewidth]{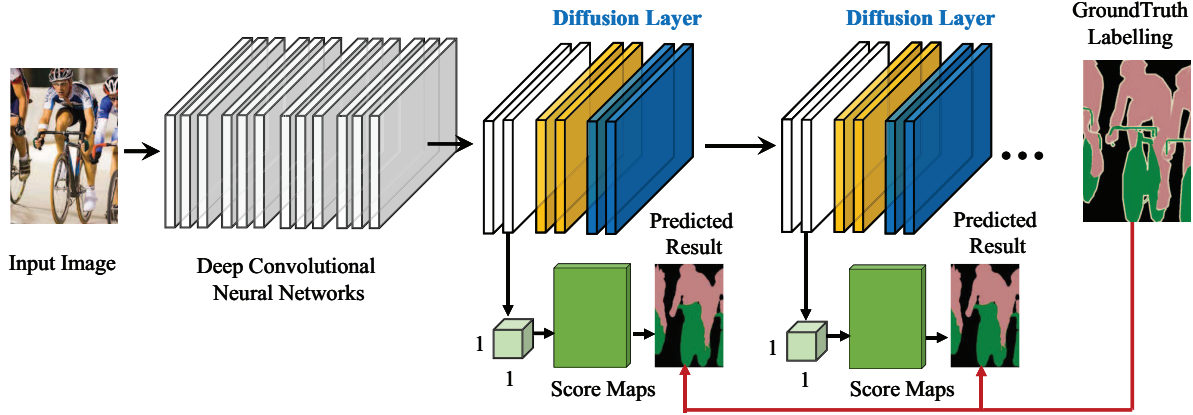}
\caption{The framework of proposed Progressively Diffused Networks for semantic segmentation. The diffusion module with several diffusion layers is stacked on the top of Deep Convolutional Neural Networks for context modeling. Each diffusion layer outputs several contextual feature maps (i.e. yellow and blue maps) for broadcasting neighborhood information on the image plane in the next layer. And depth feature maps (i.e. white maps) are also generated to communicate information of each site from one layer to the next. For the model training, the depth feature maps in each layer are convolved with $1\times1$ convolutional filters to generate the score maps for cost calculation. In the test phase, all of the score maps from distinct diffusion layers will be merged to predict the final result.}
\label{fig:framework}
\vspace{-1em}
\end{figure*}


 The performance of computer vision tasks is heavily dependent on the choice of visual representation. For that reason, many of previous efforts in deploying computer vision models focused on designing the pipelines to extract the effective visual representation~\cite{C:SparseCoding,J:DPM,C:HierRepre}. Such feature engineering based methods are important but require a lot of domain knowledge, severely limiting the development of visual applications. In order to make the vision models less dependent on feature engineer, representation learning~\cite{J:RepreLearning,J:DeepLearning}, which facilitates useful information extraction from raw data for building predictors, has attracted much attention in the past decade.

 A typical representation learning model is Convolutional Neural Networks (CNN)~\cite{J:CNN,C:ImageNet,C:FCN}, which is designed to process the data with multiple arrays such as images~\cite{C:ImageNet} or videos~\cite{J:3D-ConV}. By stacking several convolution-pooling layers, this model transforms the visual representation from one level into a slightly more abstract level. Recently, many works tend to enhance representational power of CNN by increasing the depth of architectures~\cite{C:VeryDeep,C:GoingDeeper,A:highway,C:ResNet}, and achieve great success on image classification~\cite{C:ImageNet,C:ResNet}. The dense prediction task, such as semantic segmentation, has also benefited from such deep feature learning methods~\cite{C:FCN,C:DeepLab}.  In~\cite{C:FCN}, Long \textit{et al}. firstly replaced fully-connected layers of CNN with convolutional layers, making it possible to accomplish pixel-wise prediction in the whole image by the deep model. Chen \textit{et al}.~\cite{A:deeplab2} further proposed the atrous convolution to explicitly control the resolution of feature responses, and exhibited the atrous  spatial pyramid pooling for dense predicting at multiple scales.


Meanwhile, in order to explicitly discover the intricate structures in the visual data for dense labeling, the graphic models were applied to explore the rich information (e.g. long-range dependencies or high-order potentials) in the image by defining the spatial constrains. In~\cite{C:DeepLab}, the confidence maps generated by the Fully Convolutional Networks (FCN)~\cite{C:FCN} were fed into the Conditional Random Field (CRF) with simple pairwise potentials for post-processing, but this model treated the FCN and CRF as separated components, limiting the joint optimization of the model. In contrast, Schwing \textit{et al}.~\cite{A:StructuredNetworks} jointly train the FCN and Markov Random Field (MRF) by passing the error generated by MRF back to the neural networks. However, the iterative inference algorithm (i.e. Mean Field inference) used in this method is time consuming. To improve computational efficiency, Liu \textit{et al}.~\cite{C:DPN} solve MRF by the convolution operations, which devises the additional layers to approximate the mean field inference for pairwise terms. Although these methods significantly improve the performance of dense labelling, the contextual information is still not explicitly encoded into the pixel-wise representations.

In the literature, the Long Short Term Memory (LSTM) Network has been introduced to deal with the long-range dependencies in the representation modeling, and this advanced Recurrent Neural Network (RNN) has achieved great success in many intelligent tasks~\cite{C:HandwritingLSTM,C:SequencetoSequence,C:CaptionGeneration,C:ConvLSTM}. In recent years, it has been extended to multi-dimensional communication~\cite{A:Grid-LSTM,C:SpatialLSTM,J:Hie-RNN} and adapted to represent the rich contexts in image spatial~\cite{C:LG-LSTM,C:Graph-LSTM}. In~\cite{C:LG-LSTM}, a recent advance in LSTM-based context modeling was achieved by considering both short dependencies from local area and long-distance global information from the whole image. Liang \textit{et al}.~\cite{C:Graph-LSTM} further extended this work from multi-dimensional data to general graph-structured data, and constructed an adaptive graph topology to propagate contextual information between adjacent superpixels. Nevertheless, in these works, the feature representation of each position is affected by a limited local factors (i.e. the adjacent positions), which restricts the capacity of involving diverse visual correlations in a large range. Different from using limited local LSTM units, the proposed PDNs captures the short-range and long-range dependencies from various neighbors and can generate more informative representation for pixel-wise prediction.


\section{Progressively Diffused Network}

In this section, we will first review the framework of proposed PDNs and introduce the diffusion mechanism for context modeling. The implementation details of diffusion layer will be described at the end of section.

\subsection{Framework Overview}

We develop a novel Progressively Diffused Networks (PDNs) for image context modeling, and pursue the semantic segmentation task. The architecture of proposed network is presented in Fig.~\ref{fig:framework}. The input image is first forwarded through the Deep Convolutional Neural Networks (e.g. ResNet-101~\cite{C:ResNet}) to generate a set of feature maps. Then these feature maps are fed into a series of diffusion layers to progressively spread the context information on the image plane.  After each diffused LSTM layer, the generated depth maps (i.e. white maps in Fig.~\ref{fig:framework}) are convolved with $1\times1$ filters to calculate the score maps for dense prediction. For the model training, the cross-entropy loss over all pixels is used for every diffused LSTM layer. In the test phase, the final result is predicted by merging all of the score maps from the distinct diffusion layers.






\begin{figure}[t]
\centering
\includegraphics[width=3.2in]{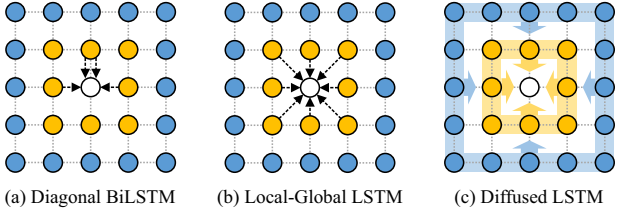}
\caption{The comparison of different types of LSTM unit. Sub-figure (a) and (b) show the previous pixel-wise LSTM unit ( i.e. Diagonal BiLSTM~\cite{C:PixelRNN} and LG-LSTM\cite{C:LG-LSTM} ) that update the states of each site by adopting fixed local factors ( i.e. adjacent sites). In (c), our proposed LSTM unit can capture the short-range and long-range dependencies from the divers neighbors and can generate more informative data representation}
\label{fig:propagation}
\vspace{-1em}
\end{figure}

\subsection{Diffusion Mechanism}
\label{D-Layer}

\begin{figure}[t]
\centering
\includegraphics[width=3in]{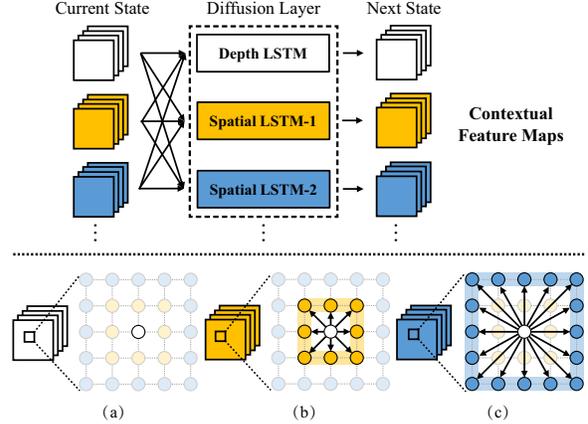}
\caption{The proposed diffused LSTM layer. \textbf{The top row} shows adopting the diffused LSTM to generate contextual feature maps.  The input of each LSTM unit is a set of contextual feature maps in current state, while the output is a special type of contextual feature maps for next state. \textbf{The bottom row} illustrates the meaning of each type of contextual feature maps. The white maps in (a) are the output of Depth LSTM, and each site indicates the information passed from the corresponding site in previous state. The yellow maps in (b) are generated by Spatial LSTM-1, and each site contains the information that the site will pass to its closest 8 neighbors in the current state. Similarly, each site in the blue maps in (c) denotes the information spreading to its second closest 16 neighbors. Best viewed in color.}
\label{fig:diffused}
\end{figure}

The diffusion mechanism aims at broadcasting the contextual information on the image plane and exploiting such information to increase the discrimination of feature representation for each pixel. In fact, such mechanism is partially inspired by biological research that the pheromone released by a single cell can affect larger tissue areas, not just the adjacent cells~\cite{J:cells}. In this article, the diffusion mechanism is adopted to spread information from one site to a large field of neighbors. It includes two coherent aspects, (1) using the information from different neighbors to enrich the feature representation of one certain site, illustrated in Fig.~\ref{fig:propagation}; (2) propagating different information from the certain site to its different neighbors to guide their further representation, illustrated in Fig.~\ref{fig:diffused}.

The diffusion mechanism in this paper is similar to recent image processing works based on LSTM~\cite{A:Grid-LSTM,C:SpatialLSTM,C:PixelRNN,C:LG-LSTM}. However, in these works, the contextual information of each position is determined by a fixed factorization (e.g. $2$ to $8$ neighboring positions), as shown in Fig.~\ref{fig:propagation} (a) and Fig.~\ref{fig:propagation} (b). Different from these locally fixed LSTM units, the modified LSTM in our PDNs allows each location to receive messages from different numbers of neighbors, as illustrated in Fig.~\ref{fig:propagation} (c). For most of the previous approaches~\cite{A:Grid-LSTM,C:LG-LSTM}, the parameters of each LSTM are shared, thus the information that each site passes to all of its neighbors is equivalent. Therefore, these methods can be viewed as a special case of proposed diffusion mechanism.


\subsection{Diffusion Layer}
\label{D-LSTM}

In practise, the diffusion layer exploits the multi-dimensional convolutional LSTMs to receive and broadcast the information. With these convolutional LSTMs, the prediction of site $\alpha$ is affected by different types of neighbors (e.g. 8 closest neighbors or 16 second closest neighbors). Thus the hidden cells of site $\alpha$ comprise the hidden cells passed from different types of neighbors and from $\alpha$ itself in the previous state. As illustrated in Fig.~\ref{fig:diffused}, we feed all of these hidden cells (denoted by contextual feature maps in Fig.~\ref{fig:diffused}) into modified convolutional LSTMs to produce individual hidden cells to $\alpha$ itself and to its different neighbors in the next state. In the following, we will describe the implementation details of diffusion layer. To be self-contained, we first recall the notation of convolutional LSTM~\cite{C:ConvLSTM} and then describe the proposed diffused LSTM.

\subsubsection{Convolutional LSTM}

Let $\mathcal{X}_t \in \mathbb{R}^{M\times N\times D}$, $\mathcal{H}_t \in \mathbb{R}^{M\times N\times D}$ and $\mathcal{M}_t \in \mathbb{R}^{M\times N\times D}$ indicate the feature maps, hidden cells and memory cells of state $t$, where $M\times N$ denotes the spatial dimension of state $t$ and $D$ is the number of channels. Similar to the traditional LSTM~\cite{A:Grid-LSTM}, we use $g^i$, $g^f$, $g^c$ and $g^o$ to indicate input gate, forget gate, memory gate and output gate respectively. And $W^i$, $W^f$, $W^c$ and $W^o$ are the gate kernels. Convolutional LSTM accepts the state $t$ as the input and outputs the next state as follows:
\begin{equation}\label{Eq:C-lstm}
\begin{split}
& g^i = \sigma( W^i_x* \mathcal{X}_t + W^i_h* \mathcal{H}_t + b^i )  \\
& g^f = \sigma( W^f_x* \mathcal{X}_t + W^f_h* \mathcal{H}_t + b^f )  \\
& g^c = \sigma( W^c_x* \mathcal{X}_t + W^c_h* \mathcal{H}_t + b^c )    \\
& g^o = \tanh( W^o_x* \mathcal{X}_t + W^o_h* \mathcal{H}_t + b^o )    \\
& \mathcal{M}_{t+1} = g^f\odot \mathcal{M}_t + g^i\odot g^c \\
& \mathcal{H}_{t+1} = g^o\odot \tanh( \mathcal{M}_{t+1})
\end{split}
\end{equation}
where $*$ denotes the convolution operator and $\odot$ denotes the Hadamard product. The symbol $\sigma$ indicates the sigmoid function. Let $\textbf{W}$ and $\textbf{B}$ indicate the collection of all gate weights and biases respectively. Following~\cite{A:Grid-LSTM} and ~\cite{C:LG-LSTM}, we define function LSTM($\cdot$) to shorten Eqn.~\eqref{Eq:C-lstm} with the form,
\begin{equation}\label{Eq:C-lstm-short}
(\mathcal{H}_{t+1},\mathcal{M}_{t+1}) = \mbox{LSTM}(\mathcal{X}_{t},\mathcal{H}_{t},\mathcal{M}_{t},\textbf{W},\textbf{B})
\end{equation}
The mechanism acts as a memory system, which records the previous information into the memory cells and is used to communicate with subsequent input. In order to ensure the next state has the same spatial dimension as the input, we need to perform zero-padding before applying the convolution operation.


\subsubsection{Diffused LSTM}

\begin{figure}[t]
\centering
\includegraphics[width=3in]{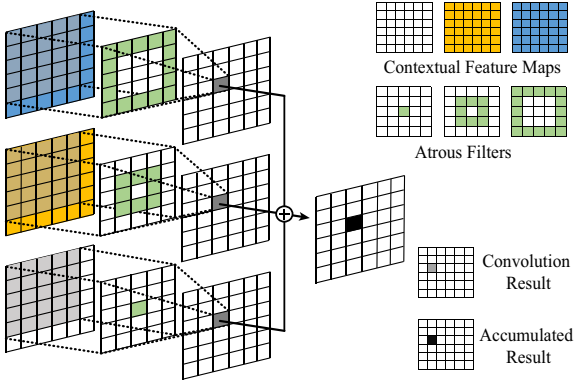}
\caption{An illustration of convolutional operations in each LSTM unit. A set of contextual feature maps with different meanings (e.g. each site in white, yellow and blue map involve the information pass to self, closest 8 neighbors and second closest 16 neighbors respectively) are fed into LSTM unit. A special type of atrous filters are designed to capture the short range and long
range dependencies from various neighbors to a certain site and pass the accumulated information to the next state. For each kernel, the green regions are learnable and others are always set to zero. }
\label{fig:convolution}
\vspace{-1em}
\end{figure}

Following the definition in~\cite{C:LG-LSTM}, we define two kinds of diffused LSTM in the proposed diffusion layer, named depth LSTM and spatial LSTM. Intuitively, the depth LSTM can maintain the previous information at each site by applying the memory cells benefited from the LSTM mechanism. The spatial LSTM calculates the information that each position travels outward to their neighbors. Note that different spatial LSTM will compute the information that each site passes to the neighbors with different spatial distances. For example, the yellow maps in Fig.~\ref{fig:diffused} are the outputs of spatial LSTM-1, and the value in each position denotes the information that the position propagates to its $3\times 3-1=8$ nearest neighbors with distance $1$. Analogously, the blue maps are the outputs of spatial LSTM-2, and the value in each position denotes the message passing to the $5\times5-3\times3=16$ second nearest neighbors with distance $2$.

As illustrated in Fig.~\ref{fig:diffused}, the input image is corresponding to $E+1$ groups of hidden cell maps (we set $E=2$ in this article for illustration), which are generated by one depth LSTM and $E$ spatial LSTM. Let $\mathcal{H}_{t,e}^s \in \mathbb{R}^{M\times N\times D}, e\in\{1,2,...,E\}$ denote the $e$-th group of hidden cell maps generated from $e$-th spatial LSTM, and the hidden cells in each position are used to propagate the information to its $(2e+1)^2-(2e-1)^2=8e$ neighbors with distance $e$. Let $\mathcal{H}_{t}^d \in \mathbb{R}^{M\times N\times D}$ indicate the hidden cell maps calculated by the depth LSTM using the weights updated in the $t$-th layer. Thus
the gate values of a certain LSTM unit (e.g. depth LSTM or spatial LSTM) in $t$-th layer can be calculated by,
\begin{equation}\label{Eq:our-lstm}
\begin{split}
& g_t^i = \sigma ( \sum\nolimits_e \{W_{t,e}^s\}^i * \mathcal{H}_{t,e}^s + \{W_{t}^d\}^i * \mathcal{H}_{t}^d + b_t^i )  \\
& g_t^f = \sigma ( \sum\nolimits_e \{W_{t,e}^s\}^f * \mathcal{H}_{t,e}^s + \{W_{t}^d\}^f * \mathcal{H}_{t}^d + b_t^f )  \\
& g_t^c = \sigma ( \sum\nolimits_e \{W_{t,e}^s\}^c * \mathcal{H}_{t,e}^s + \{W_{t}^d\}^c * \mathcal{H}_{t}^d + b_t^c )  \\
& g_t^o \!=\! \tanh ( \sum\nolimits_e \{W_{t,e}^s\}^o * \mathcal{H}_{t,e}^s + \{W_{t}^d\}^o * \mathcal{H}_{t}^d + b_t^o )  \\
\end{split}
\end{equation}
where $W_{t,e}^s$ and $W_{t}^d$ indicate the weights of
kernels associated with $e$-th spatial hidden cell maps and depth hidden cell maps in $t$-th layer. And the superscript $i$, $f$, $c$ and $o$ correspond to distinct state gates.

Denote the $e$-th group of memory cells for the spatial dimension as $\mathcal{M}_{t,e}^s \in \mathbb{R}^{M\times N\times D}$ and the memory cells for depth dimension as $\mathcal{M}_{t}^d \in \mathbb{R}^{M\times N\times D}$. Same as convolutional LSTM, the novel hidden cell maps and memory cell maps in $t+1$-th layer are computed as,
\begin{equation}\label{Eq:diffused_layer}
\begin{split}
(\mathcal{H}_{t+1,1}^s,\mathcal{M}_{t+1,1}^s) & \!=\! \mbox{LSTM}(\mathcal{H}_{t}^\dag,\mathcal{H}_{t}^d,\mathcal{M}_{t,1}^s,\{\textbf{P}\}_{t,1}^s) \\
(\mathcal{H}_{t+1,2}^s,\mathcal{M}_{t+1,2}^s) & \!=\! \mbox{LSTM}(\mathcal{H}_{t}^\dag,\mathcal{H}_{t}^d,\mathcal{M}_{t,2}^s,\{\textbf{P}\}_{t,2}^s)
\\
&... \\
(\mathcal{H}_{t+1,E}^s,\mathcal{M}_{t+1,E}^s) & \!=\! \mbox{LSTM}(\mathcal{H}_{t}^\dag,\mathcal{H}_{t}^d,\mathcal{M}_{t,E}^s,\{\textbf{P}\}_{t,E}^s)
\\
(\mathcal{H}_{t+1}^d,\mathcal{M}_{t+1}^d) & \!=\! \mbox{LSTM}(\mathcal{H}_{t}^\dag,\mathcal{H}_{t}^d,\mathcal{M}_{t}^d,\{\textbf{P}\}_{t}^d)
\\
\end{split}
\end{equation}
where $\mathcal{H}_{t}^\dag = \{\mathcal{H}_{t,e}^s\}_{e=1}^E$ is the set of spatial hidden cell maps. $\textbf{P}=\{\textbf{W},\textbf{B}\}$ indicates the parameter set.

In the above process, different numbers of spatial LSTMs allow us arbitrarily enlarge \textit{field-of-view} in the context modeling. For a LSTM unit in the certain layer, there exist $E+1$ filters with distinct forms, and each one is associated with a group of hidden cell maps. As illustrated in Fig.~\ref{fig:convolution}, we design a novel type of atrous filters~\cite{A:deeplab2} to calculate the convolutional results. Thus depth hidden cell maps specify the depth filter and only the center of the kernel has weight value. The $e$-th spatial filter is associated with $e$-th group of spatial hidden cell maps, and it introduces non-zero weights in the sites whose distance to the kernel center is $e$. Note that, if the site $\alpha$ has neighbor $\alpha'$ with distance $e$, we need to adopt the hidden cells of site $\alpha'$ in the $e$-th group to enrich the representation of site $\alpha$.

In this way, each site in the input image can provide distinct guidance to its neighbors with different distances by employing specific spatial LSTMs, which takes the spatial layouts and interactions into account for feature learning. In order to ensure different neighbors receive various information, the weight matrices $\textbf{W}_{t}^s$ and bias  $\textbf{B}_{t}^s$ of $E$ spatial LSTMs are not shared in this article.

\section{Experiments}

In this section, we demonstrate the  effectiveness of proposed PDNs by comparing it with state-of-the-art semantic segmentation methods. In the following, we first give a brief overview of the datasets and evaluation metrics. Then we report the performance of PDNs on both object segmentation and object part segmentation tasks. The component analysis is also investigated at the end of this section.

\textbf{Datasets and Evaluation Metrics}. We compare proposed PDNs with the state-of-the-art methods on ImageNet Parsing, PASCAL-Person-Part and PASCAL VOC 2012 (VOC2012) datasets. \textbf{ImageNet Parsing}~\cite{A:ImageNetParsing} is a challenging dataset for scene-centric semantic segmentation task. It includes 150 semantic categories, and many categories have the similar appearance, prompting us to use more reasonable contextual reasoning for dense prediction. In this dataset, 20,210 images are employed for model training and another 2,000 images for validation. \textbf{PASCAL-Person-Part} dataset is a fine-grained part segmentaion benchmark collected by Chen \textit{et. al}.~\cite{C:Pascal-Part} from PASCAL VOC 2010 dataset. It contains the detailed part annotation for each person and the annotations are merged into six person parts ( i.e. Head, Torso, Upper/Lower Arms and Upper/Lower Legs ) and one background category as previous setting in~\cite{A:zoom,A:Scale-aware,C:Graph-LSTM}. Totally, 1,716 images are used for model training and 1,817 for test. We also report the performance on the \textbf{PASCAL VOC2012}~\cite{J:PascalVOC} dataset, since it is a standard benchmark for generic semantic segmentation. It contains 20 object categories and the sizes of training, validation and test set are 10582, 1449 and 1456, respectively. For all of the above datasets, we employ standard mean intersection-over-union criterion (denoted as mIoU) and pixel-wise accuracy~\cite{C:FCN} to evaluate our proposed architecture and other comparative methods.

\begin{figure*}[t]
\centering
\includegraphics[width=\linewidth]{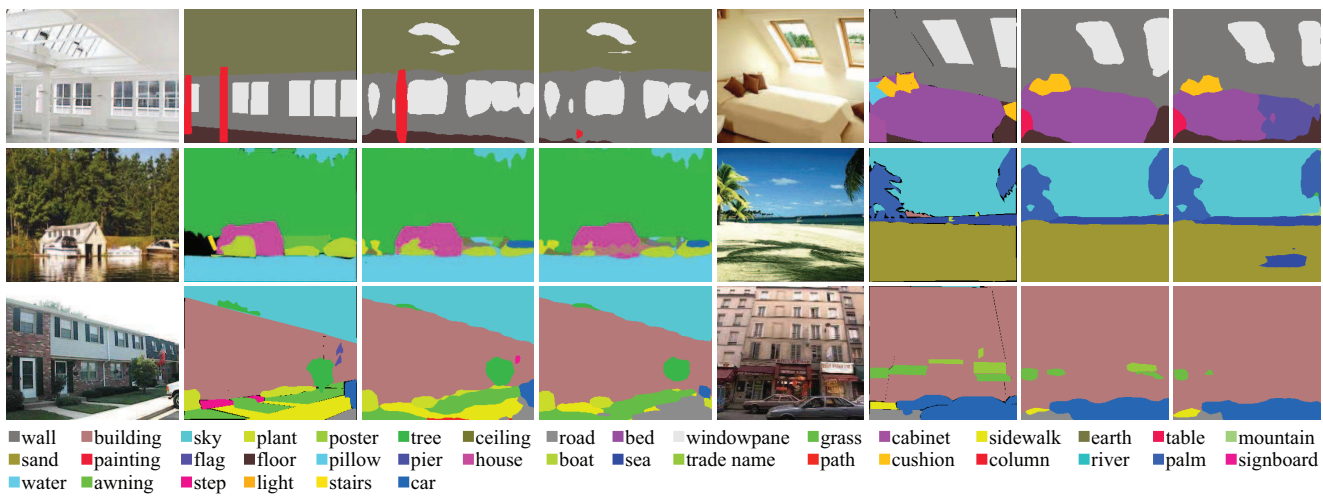}
\caption{ Visualization of segmentation results on ImageNet Parsing dataset~\cite{A:ImageNetParsing}. From left to right are input image, groundtruth labeling, segmentation result by proposed PDNs and segmentation by ResNet-101~\cite{C:ResNet}. Best viewed in color.}
\label{fig:Imagenet}
\vspace{-0.5em}
\end{figure*}

\begin{figure*}[th]
\centering
\includegraphics[width=\linewidth]{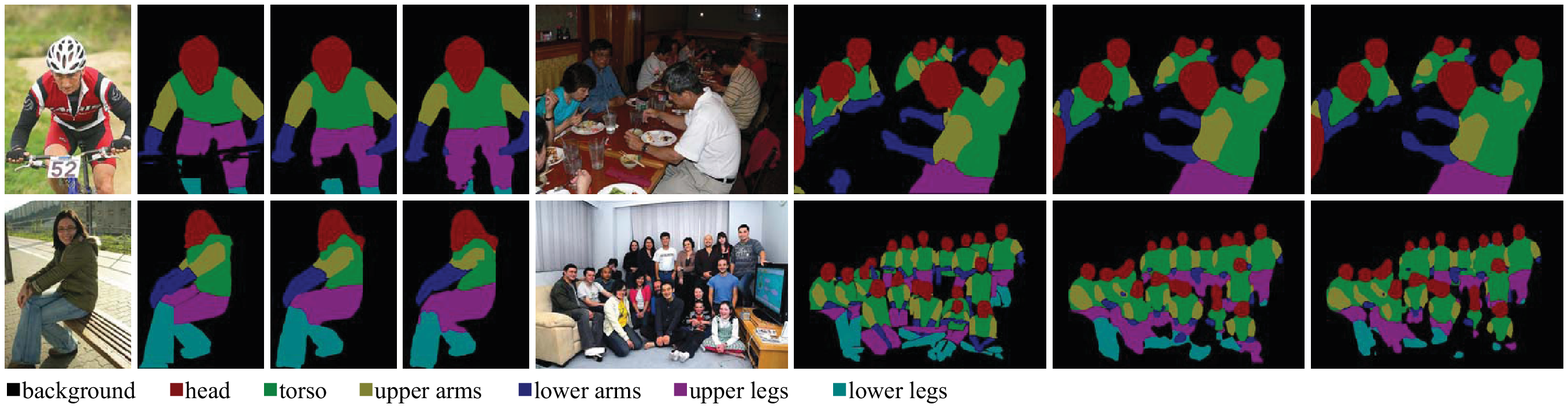}
\caption{ Visualization of segmentation results on PASCAL-Person-Part dataset~\cite{C:Pascal-Part}. From left to right are input image, groundtruth labeling, segmentation result by proposed PDNs and segmentation by ResNet-101~\cite{C:ResNet}. Best viewed in color.}
\label{fig:person}
\vspace{-0.5em}
\end{figure*}

\begin{figure*}[th!]
\centering
\includegraphics[width=\linewidth]{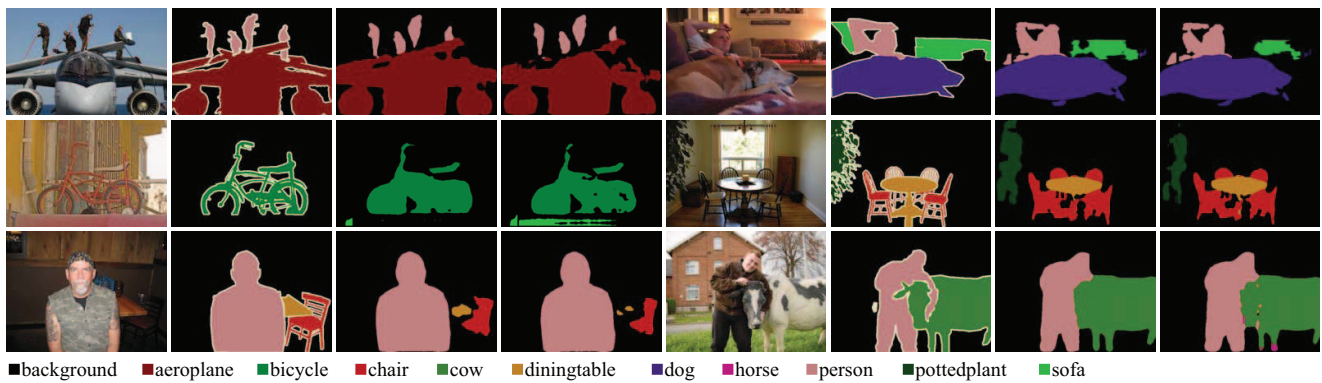}
\caption{ Visualization of segmentation results on PASCAL VOC2012 dataset~\cite{J:PascalVOC}. From left to right are input image, groundtruth labeling, segmentation result by proposed PDNs and segmentation by ResNet-101~\cite{C:ResNet}. Best viewed in color. }
\label{fig:voc}
\vspace{-0.5em}
\end{figure*}

\textbf{Implementation Details}. In our experiments, our architecture is implemented based on Caffe platform~\cite{C:Caffe} and all networks are trained on four NVIDIA GeForce GTX TITAN X GPUs with 12GB memory. The input image is randomly cropped to $321\times321$ for model training. Five diffusion layers are stacked onto top of the convolutional neural networks (i.e. ResNet-101\footnote{\url{http://liangchiehchen.com/projects/DeepLab.html}}) and we fine-tune the network based on the pre-trained neural network in the bottom. The learning rate of the newly added layers is initialized as $2.5 \times 10^{-3}$ and that of other previously learned layers is initialized as $2.5 \times 10^{-4}$. All the parameters in the diffusion layer are randomly initialized from a Gaussian distribution with the mean 0 and the variance 0.01.  We train all the models using stochastic gradient descent with a batch size of 1 image, momentum of 0.9, and weight decay of 0.0005.

\begin{table}[t]
\begin{center}
\begin{tabular}{c|c|c}

\hline
Method & pixel acc. \% & mean IoU \%\\
\hline
FCN~\cite{C:FCN}          &69.05 & 24.86  \\
DeepLab~\cite{C:DeepLab}  &71.06 & 27.08 \\
Grid-LSTM~\cite{A:Grid-LSTM}  &69.37& 25.11\\
LG-LSTM~\cite{C:LG-LSTM}  &69.91&  25.79\\
ResNet~\cite{C:ResNet}    &75.09 & 35.07  \\
ResNet + CRF~\cite{C:ResNet}    &77.22 & 36.79 \\
\hline
\hline
PDNs &\textbf{77.51} &\textbf{37.02}  \\
\hline
\end{tabular}
\end{center}
\caption{Experimental results on ImageNet Parsing \textit{val} set.}
\label{tbl:Result_ImageNet_test}
\end{table}


\begin{table*}[t]
\begin{center}
\resizebox{5.2in}{!}{%
\begin{tabular}{c|c|c|c|c|c|c|c|c}
\hline
Method & head & torso & u-arms & l-arms & u-legs & l-legs& background &\textbf{mIoU} \\
\hline
DeepLab~\cite{C:DeepLab}       &78.1& 54.0 &37.3&36.9& 33.7&29.6&92.9&51.8 \\
HAZN~\cite{A:zoom}              &80.8&59.1  &43.1&42.8& 39.0&34.5&93.6& 56.1\\
Attention~\cite{A:Scale-aware} &-&-  &-&-&- &-&-&56.4 \\
Grid-LSTM~\cite{A:Grid-LSTM}   &81.9& 58.9 &43.1&46.9&40.1 &34.6&86.0&56.0 \\
LG-LSTM~\cite{C:LG-LSTM}       &82.7& 61.0 &45.4&47.8&42.3 &38.0&88.6&58.0 \\
Graph-LSTM~\cite{C:Graph-LSTM}  &82.7& 62.7 &46.9&47.7& 45.7&40.9&94.6&60.2 \\
ResNet~\cite{C:ResNet}    &85.1& 68.4 &50.9&51.2& 46.7&39.4&\textbf{95.6}& 62.4\\

ResNet + CRF~\cite{C:ResNet}    &84.2 &68.8&51.2&52.2 &46.8&39.4& 95.3& 62.6\\

\hline

PDNs & \textbf{85.6} &\textbf{70.0}&\textbf{53.3}&\textbf{53.2} &\textbf{47.5}&\textbf{43.7}&\textbf{95.6}&\textbf{64.1}
 \\

\hline
\end{tabular}}
\end{center}
\caption{Experimental results (\textbf{IoU}) on PASCAL-Person-Part set.}
\label{tbl:BreakDown_VOC_part}
\end{table*}

\begin{table*}[t]\renewcommand{\arraystretch}{1.2}
\begin{center}
\resizebox{\textwidth}{!}{%
\begin{tabular}{c|c|c|c|c|c|c|c|c|c|c|c|c|c|c|c|c|c|c|c|c|c}

\hline
Method  & aero & bike & bird & boat& bottle& bus & car & cat &chair & cow & table & dog & horse & mbike & person & plant & sheep& sofa & train & tv  & \textbf{mIoU} \\
\hline
FCN~\cite{C:FCN}              & - &-&-&-&-&-&-&-&-&-&-&-&-&-&-&-&-&-&-&-& 62.7 \\
DeepLab~\cite{C:DeepLab}      &78.7&34.3&73.8&61.5&67.2&83.7&78.2&80.9&29.7 &74.0&52.1&72.2&68.4&73.6&80.6&48.4&72.4&43.6&77.0&59.1& 66.7 \\
CRF-RNN~\cite{C:RNN}          & - &-&-&-&-&-&-&-&-&-&-&-&-&-&-&-&-&-&-&-& 69.6  \\
DPN~\cite{C:DPN}          & 84.8  & 37.5 & 80.7& 66.3& 67.5 & 84.2& 76.4& 81.5& 33.8& 65.8& 50.4& 76.8&67.1&74.9& 81.1&48.3&75.9&41.8&76.6&60.4& 67.8  \\
LG-LSTM$^\dagger$~\cite{C:LG-LSTM}      & 77.3 &29.7&74.7&58.4&60.6&78.2&76.9& 79.2&26.5&51.7&53.6&73.9&62.4&68.1&75.9&43.4&60.7&42.5&78.1&56.7&62.8  \\
ResNet$^\dagger$~\cite{C:ResNet}        & \textbf{87.9} &41.4&\textbf{89.5}&72.5&\textbf{80.7}&93.0&87.7 &\textbf{91.7}&\textbf{39.7}&83.2&\textbf{53.8}& \textbf{85.0}&\textbf{85.2}&82.5&85.6&59.8&85.5&40.2&87.0&77.2&76.3  \\

\hline

PDNs$^\dagger$-NB24  & 87.7&42.8&89.4&\textbf{73.4}&80.4&93.1 &88.8&91.4&39.1&83.7&51.7&84.7&84.8 &83.6&86.1&\textbf{60.4}&\textbf{87.1}&\textbf{42.7}&\textbf{87.4}&78.2& \textbf{76.7} \\

PDNs$^\dagger$-NB48 &87.2 &\textbf{43.5}&89.2&71.2&78.6& \textbf{93.4}&\textbf{88.9}&91.1&39.5&\textbf{83.8}& 47.6&84.8&83.0&\textbf{83.9}&\textbf{86.9}&60.1&86.8&42.6&86.5&\textbf{79.0}&76.4 \\

\hline
\end{tabular}}%
\end{center}
\caption{Experimental results (\textbf{IoU}) on PASCAL VOC 2012 \textit{val} set. The approaches pre-trained on COCO~\cite{C:COCO} are marked with $^\dagger$.}
\label{tbl:BreakDown_VOC12_val}
\end{table*}

\subsection{Experiment Results and Comparisons}

In the following, we compare the proposed Progressively Diffused Networks with several state-of-the-art methods on image segmentation tasks.

\textbf{ImageNet Parsing dataset}~\cite{A:ImageNetParsing}. We compare our model with five state-of-the-art methods: FCN~\cite{C:FCN}, DeepLab~\cite{C:DeepLab}, Grid-LSTM~\cite{A:Grid-LSTM}, LG-LSTM~\cite{C:LG-LSTM} and ResNet~\cite{C:ResNet}. These methods can be grouped into two categories: i) CNN-based method: FCN~\cite{C:FCN}, DeepLab~\cite{C:DeepLab} and ResNet~\cite{C:ResNet}; ii) LSTM-based method: Grid-LSTM~\cite{A:Grid-LSTM}, LG-LSTM~\cite{C:LG-LSTM}. The first category employs convolution and pooling operations to directly extract the abstract feature representation. The second category uses LSTM to construct short-distance and long-distance spatial dependencies.

Since ImageNet Parsing is a newly proposed large-scale segmentation dataset, and previous works didn't report their performance on it, all the results in Table~\ref{tbl:Result_ImageNet_test} are given based on our own implementation. All of the comparison methods have trained 200,000 iterations without any extra data. We record the results every ten thousand iterations and select the best to report. According to Table~\ref{tbl:Result_ImageNet_test}, the proposed PDNs outperforms the baseline method ResNet-101~\cite{C:ResNet} in terms of pixel accuracy and mean IoU with 2.42\% and 1.95\% respectively. It well demonstrates that incorporating the contextual information into representation learning can further enhance the discriminative ability of deep features. Fig.~\ref{fig:Imagenet} gives the visualization results of our method and ResNet-101~\cite{C:ResNet}. It is obvious that the proposed PDNs can effectively distinguish similar objects in complex scenes by local and global reasoning.

\textbf{PASCAL-Person-Part dataset}~\cite{C:Pascal-Part}. Table~\ref{tbl:BreakDown_VOC_part} shows the comparison results with seven state-of-the-art approaches~\cite{C:DeepLab,A:zoom,A:Scale-aware,A:Grid-LSTM,C:LG-LSTM,C:Graph-LSTM,C:ResNet} on the metric of mean IoU. An obvious improvement, i.e. 1.7\% increased by PDNs over the ResNet-101, can be observed from the comparison on breakdown categories. In order to reflect the advantages of proposed PDNs in context modeling, we also add CRF~\cite{C:DeepLab} on the top of ResNet-101~\cite{C:ResNet} to optimize prediction results. Such model achieves mean IoU of 62.6\%, which is still 1.5\% less than our PDNs. The visualization of segmentation results on this dataset are shown in Fig.~\ref{fig:person}. Our PDNs gives more consistent segmentation results by incorporating spatial dependencies.

\textbf{PASCAL VOC 2012 dataset}~\cite{J:PascalVOC}. We report the segmentation results and comparisons with six recent state-of-the-arts~\cite{C:FCN,C:DeepLab,C:RNN,C:DPN,C:LG-LSTM,C:ResNet} in VOC2012. Table~\ref{tbl:BreakDown_VOC12_val} shows the breakdown performance of proposed PDNs and comparisons with six state-of-the-arts on VOC2012 \textit{val} dataset. Generally, the proposed PDNs outperforms other methods on vast majority of categories. Compared with the above two datasets, our method doesn't achieve a better improvement than ResNet~\cite{C:ResNet}. One reasonable explanation is that the benefit of context modeling degrades at such dataset. This is because scenes in VOC2012 are less complex than ImageNet Parsing~\cite{A:ImageNetParsing} (e.g. less than three foreground objects in an image, obvious differences in objects appearance), and preserving the boundaries of objects is more beneficial than global reasoning. However, in terms of predicting semantic categories with the dramatic changes in appearance and size such as plant, sheep and sofa, our method still achieves a improvement. Fig.~\ref{fig:voc} gives some exemplar segmentation results on VOC2012.

\subsection{Comparison to Different Variants}



To strictly evaluate the effectiveness of different numbers of spatial neighbors, we also report the performance of using 24 spatial neighbors and 48 spatial neighbors. We use the capital ``NB" to indicate the number of spatial neighboring connections for each pixel. According to Table~\ref{tbl:BreakDown_VOC12_val}, ``PDNs-NB24" with 24 spatial neighbors outperforms that with 48 spatial neighbors by 0.3\% over the metric of mIoU. Intuitively, too many neighbors, while bringing more contextual information, also generate much noise, the latter affecting the final prediction results.


We also validate the advantage of using different diffusion layers on the top of ResNet-101~\cite{C:ResNet} on PASCAL-Person-Part~\cite{C:Pascal-Part} dataset. Table~\ref{tbl:Self_VOC_part} shows the breakdown IoU results, and the improvements can be observed by
gradually using more diffusion layers. It verifies well the
effectiveness of exploiting more discriminative feature representation by stacking multi-layers to diffuse the information. At the same time, we can find that the performance change between the fourth layer and the fifth layer is small. Intuitively, we can assume that the information has been effectively propagated in the image plane in the fifth layer.

\begin{table}[ht]
\begin{center}
\begin{tabular}{c|c|c}
\hline
Method & pixel acc. \% &mean IoU \% \\
\hline

PDNs-1  &93.7 & 63.2\\
PDNs-2  &93.8 &63.6\\
PDNs-3  &93.8    &63.6\\
PDNs-4  &\textbf{93.9}    &\textbf{64.0} \\
PDNs-5  &93.8&63.9\\
\hline
\hline
PDNs-merge  &\textbf{93.9} &\textbf{64.1} \\

\hline
\end{tabular}
\end{center}
\caption{Experimental results with different numbers of diffusion layers on PASCAL-Person-Part set.}
\label{tbl:Self_VOC_part}
\end{table}

\section{Conclusion}

This work proposes a novel progressively diffused networks (PDNs) for context modeling, and pursues the semantic segmentation task. In PDNs, the stacked diffusion layers make the contextual information spreading from a certain site to a large range on the image plane, through adopting convolutional LSTM with special atrous filters to capture the short range and long range spatial dependencies for a certain site and passing the accumulated information to site self or its neighbors in the next layer. The experiment results on three benchmarks demonstrate the effectiveness of proposed architecture.

{\small
\bibliographystyle{ieee}
\bibliography{egbib}
}

\end{document}